# Dual-Channel Latent Factor Analysis Enhanced Graph Contrastive Learning for Recommendation

Junfeng Long and Hao Wu, *Member*, *IEEE*

*Abstract*—Graph Neural Networks (GNNs) are powerful learning methods for recommender systems owing to their robustness in handling complicated user-item interactions. Recently, the integration of contrastive learning with GNNs has demonstrated remarkable performance in recommender systems to handle the issue of highly sparse user-item interaction data. Yet, some available graph contrastive learning (GCL) techniques employ stochastic augmentation, i.e., nodes or edges are randomly perturbed on the user-item bipartite graph to construct contrastive views. Such a stochastic augmentation strategy not only brings noise perturbation but also cannot utilize global collaborative signals effectively. To address it, this study proposes a latent factor analysis (LFA) enhanced GCL approach, named LFA-GCL. Our model exclusively incorporates LFA to implement the unconstrained structural refinement, thereby obtaining an augmented global collaborative graph accurately without introducing noise signals. Experiments on four public datasets show that the proposed LFA-GCL outperforms the state-of-the-art models.

## I. INTRODUCTION

Recently, graph-based representation learning has shown impressive achievements across various domains[1-16], with graph neural networks (GNNs) notably succeeding in recommender systems [17-19]. Recommender systems are commonly utilized in various areas of society, such as providing personalized information to users in online shopping [24] and social networks [25]. Additionally, they can optimize the quality of service prediction [26, 27, 47] management to enhance user experience. The mainstream solutions for recommender systems have developed from neighborhood-based collaborative filtering (CF) methods [28-31] to deep learning CF frameworks [32-34]. Among the various deep learning frameworks, the GNN-based recommendation method achieves notable results by stacking multiple layers of neighborhood aggregation to extract local collaborative signals from connected neighborhoods [35].

Although GNN-based recommendation models are effective, they typically adhere to the supervised learning paradigm [40], which requires abundant, high-quality labeled data. However, current recommender systems often lack supervised signals because users interact with only a small fraction of items, leading to sparse observable interaction data [42, 43, 46]. As a result, GNN-based recommender systems face challenges in learning optimal node representations for recommendation tasks, especially when labeled data is scarce.

To alleviate the issue of sparsity data, on the one hand, the methods based on matrix factorization [36-39], such as Latent Factor Analysis (LFA) [41, 45], have achieved good results in various tasks with sparse data [20-23] by filling missing values by abstracting graphs into sparse matrices [56-59]. On the other hand, in the GNN based, data augmentation techniques are used to solve the problem of data sparsity, such as GraphDA [55], employ pre-training and Top-$K$ sampling to acquire user-user and item-item correlation matrices, which fill the user-item interaction bipartite graph. Additionally, it is better to combine augmented data with contrastive learning. Many recent studies [48, 52] have introduced contrastive learning (CL) into GNN-based recommendation methods, called graph contrastive learning (GCL), and have shown excellent performance. GCL generates multiple views [53] of the same sample through data augmentation, maximizing the mutual information between positive samples while pushing negative samples apart. This approach is beneficial for learning more coherent user/item node representation and helps alleviate popularity bias. For example, SGL [63] employs random walk sampling and probabilistically drops nodes/edges to generate two additional augmented views for contrastive learning. DCL [60] constructs two augmented views by applying random perturbations to $L$-hop subgraphs centered around each user/item. They both employ random data augmentation on graph structures for contrastive learning. SimGCL [64] does not employ data augmentation on the graph structure. Instead, it adds Gaussian noise into node representation to build a contrastive view.

While the aforementioned GCL-based recommendation methods have demonstrated impressive results, these methods may have some limitations: a) constructing data augmentation views using completely random data perturbations may introduce unnecessary noisy data or result in the loss of important node information, which may mislead the learning of node representation; and b) the augmented views constructed by these data augmentation methods are still sparse and do not provide good global information.

To cope with the previously mentioned problems, we propose a LFA enhanced GCL approach, named LFA-GCL, which obtains the augmented graph following the principle of LFA. Such an augmentation strategy not only provides valuable global collaborative signals but also avoids noise perturbation when performing contrastive learning. The main contributions are as follows:

- We improve recommendation performance by designing an effective graph contrastive learning framework.

➢ J. F. Long is with the School of Computer Science and Technology, Chongqing University of Posts and Telecommunications, Chongqing 400065, China (e-mail: june14223@gmail.com).
➢ H. Wu is with the College of Computer and Information Science, Southwest University, Chongqing 400715, China (e-mail: haowuf@gmail.com).

- We employ a global collaborative signals learning method LFA to generate an augmented view without introducing irrelevant noise signals.
- We validate the excellent performance of the proposed LFA-GCL by conducting experiments and analysis on real-world datasets.

## II. PRELIMINARIES

### A. Problem Formulation

*Definition:* Given a graph $G = (V, E)$, let $I$ and $U$ represent the sets of items and users, where $V = I \cup U$ contains all items and users, and $E = \{e_{ui} | i \in I, u \in U\}$ represents the set of edges existing in graph $G$. An adjacency matrix $A \in \mathbf{R}^{|U| \times |I|}$ can be used to represent user-item interactions.

### B. Latent Factor Analysis Model

According to [54, 65, 66, 68, 69], an LFA model constructs a low-rank approximation of sparse data from known data. Given a matrix $R$ and its element set $\Lambda$, LFA model builds a prediction $\hat{R}=PQ^T$ through the latent factor matrices $P \in \mathbf{R}^{|U| \times f}$ and $Q \in \mathbf{R}^{|I| \times f}$, $f$ represents the dimension of the latent features, and $f \ll \min\{|U|, |I|\}$. Thus, a Euclidean distance-based objective is given as:

$$\begin{aligned}\varepsilon &= \left(\left(R-\hat{R}\right)^2 + \lambda\left(\|P\|_F^2 + \|Q\|_F^2\right)\right) \\ &= \sum_{r_{ui} \in \Lambda}\left(\left(r_{ui}-\hat{r}_{ui}\right)^2 + \left(\|p_u\|_2^2 + \|q_i\|_2^2\right)\right),\end{aligned} \quad (1)$$

where $\lambda$ represents the regularization coefficient, $r_{ui}$ and $\hat{r}_{ui}$ respectively represent an element in $R$ and $\hat{R}$, $p_u$ represents the $u$-th row vector of $P$, $q_i$ represents the $i$-th column vector of $Q$, $\|\cdot\|_F$ represents the Frobenius norm, and $\|\cdot\|_2$ represents the L2 norm [49-51, 74].

### C. Graph Convolutional Network

GCN is an effective graph representation learning method. Its core idea is to perform a neighborhood aggregation mechanism on the graph G to update the own representation of a node by aggregating the representation of its neighboring nodes in the neighborhood:

$$M^{(l)} = H(M^{(l-1)}, G), \quad (2)$$

where $M^{(l)}$ represents the representation of all nodes at the $l$-th layer, $M^{(l-1)}$ represents the representation of the $(l-1)$-th layer, and the function $H(\cdot)$ is used to aggregate neighborhood information. Typically, this process is repeated multiple times to obtain node information from high-order neighbors. From the vector level, it is clearer to subdivide it into two processes: propagation and aggregation:

$$m_u^{(l)} = f_{propagate}(\{m_v^{(l-1)} | v \in N_u\}), \quad (3)$$

$$m_u = f_{readout}([m_u^{(0)}, m_u^{(1)}, ..., m_u^{(L)}]), \quad (4)$$

where $m_u^{(0)}$ represents the initial embedding of user $u$, $N_u$ contains all of the adjacent items of user $u$ in the interaction graph $G$, and $L$ represents the number of convolutional layers. $f_{propagate}(\cdot)$ obtains the representation of user $u$ at $l$-th layer by aggregating the representation of its neighbor nodes at the previous layer. After undergoing $L$ iterations of aggregation, the information of $L$-hop neighbors for user $u$ will be aggregated into $m_u^{(L)}$. The function $f_{readout}(\cdot)$ integrates the node representation $[m_u^{(0)}, m_u^{(1)}, ..., m_u^{(L)}]$ of user $u$ at each layer in some way to obtain its final representation $m_u$. The implementation of the $f_{readout}(\cdot)$ mainly includes last-layer only [67], concatenation [72], and weighted sum [17]. Using the same method, the node representation $m_i$ for the item $I$ can be got.

### D. Graph Contrastive Learning

GCL is usually used as an auxiliary task in multi-task joint learning, which enhances recommendation by exploiting self-supervised signals [63] and maximizing the mutual information between the same nodes of different views augmented from the original graph. Typically, the GCL method uses InfoNCE [70] loss function:

$$L_{cl} = \sum_{t \in B} -\log \frac{exp(f(z_t^{'T} z_t^{''})/\tau)}{\sum_{v \in B} exp(f(z_t^{'T} z_v^{''})/\tau)}, \quad (5)$$

where $z'$ and $z''$ denote the representation learned from each of the two views, $f(\cdot)$ represents the similarity function, $B$ is the set of users or items, and $\tau$ represents the temperature parameter.

## III. METHOD

In this part, we describe the LFA-GCL model in more detail. Its framework is exemplified in Fig. 1, which contains four modules:

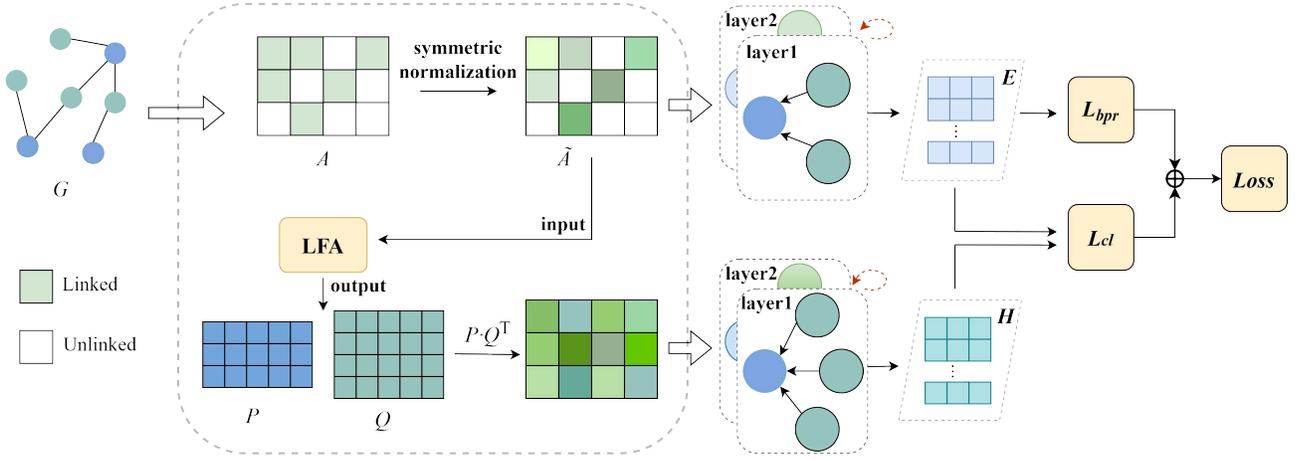

Figure 1. Performance of different user sparse interaction degree groups

- **The graph convolutional module** utilizes LightGCN [17], a simplified GCN structure, to learn the node representation of the main view (original graph).
- **The data augmentation module** constructs a view through data augmentation and uses graph convolutional module to learn the augmented node representation.
- **The graph contrastive learning module** calculates the contrastive loss for the node representation that is learned from the two different views.
- **The joint optimization module** combines recommendation task loss on the main view and the dual-view contrastive learning loss to optimize the overall performance.

*A. Graph Convolutional Module*

As mentioned in Section C of the preliminaries, GCN-based recommendation methods learn a representation for each node by propagating and aggregating messages on graph $G$. According to problem formulation, graph $G$ can be represented by the adjacency matrix $A$, and then we perform symmetry normalization on it to obtain $\tilde{A}$. We assign embedding vectors $e_k^{(u)} \in \mathbf{R}^d$ and $e_j^{(i)} \in \mathbf{R}^d$ of dimension $d$ to user $u_k$ and item $i_j$ respectively. The collections of embedding for all users and all items are marked as $E^{(u)} \in \mathbf{R}^{|U| \times d}$ and $E^{(i)} \in \mathbf{R}^{|I| \times d}$, respectively. We use a convolutional network to aggregate the information from neighboring nodes, and the convolution process at the $l$-th layer is represented as:

$$c_{k,l}^{(u)} = p\left(\tilde{A}_{k,:}\right) \cdot E_{l-1}^{(i)}, \quad c_{j,l}^{(i)} = p\left(\tilde{A}_{:,j}\right) \cdot E_{l-1}^{(u)}, \tag{6}$$

where $c_{k,l}^{(u)}$ and $c_{j,l}^{(i)}$ represent the node embedding aggregated for user $u_k$ and item $i_j$ in the $l$-th layer. We employ an edge dropout method $p(\cdot)$ to mitigate the model overfitting problem. We sum the node embeddings of each layer to obtain the final representation of each node, and use the inner product between the final representation of user $u_k$ and item $i_j$ as the preference of user $u_k$ for item $i_j$:

$$e_k^{(u)} = \sum_{l=0}^{L} c_{k,l}^{(u)}, \quad e_j^{(i)} = \sum_{l=0}^{L} c_{j,l}^{(i)}, \quad \hat{y}_{k,j} = e_k^{(u)\mathrm{T}} e_j^{(i)} \tag{7}$$

*B. Data Augmentation Module*

We adopt a pre-training approach using LFA to predict the matrix $\tilde{A}$ to alleviate the issue of sparse supervision signals, which predicts potential preference information of users for items from existing interactions[75]. We first initialize $P$ and $Q$, and then use the alternating least squares (ALS) method [71, 76] to optimize (1) to obtain $\hat{R}=PQ^{\mathrm{T}}$, and stochastic gradient descent(SGD) [77-79] algorithm can also be used here. This matrix $\hat{R}$ represents the latent preference of users for items, which not only provides rich supervision signals but also provides global collaborative signals for the model. We perform message passing and aggregation on users and items at each layer on $\hat{R}$ to learn augmented node embeddings:

$$h_{k,l}^{(u)} = \hat{R}_{k,:} \cdot E_{l-1}^{(i)}, \quad h_{j,l}^{(i)} = \hat{R}_{j,:}^{\mathrm{T}} \cdot E_{l-1}^{(u)}, \tag{8}$$

where $h_{k,l}^{(u)}$ and $h_{j,l}^{(i)}$ represent the augmented node embeddings learned for user $u_k$ and item $i_j$ at the $l$-th layer. Similarly, we also sum the node embeddings of layers to obtain the final augmented node representation:

$$h_k^{(u)} = \sum_{l=0}^{L} h_{k,l}^{(u)}, \quad h_j^{(i)} = \sum_{l=0}^{L} h_{j,l}^{(i)}. \tag{9}$$

We use $P \cdot Q^{\mathrm{T}}$ instead of $\hat{R}$ to avoid computing and storing large and dense matrices in the computation, thereby improving efficiency. We use the latent feature matrices and embedding sets to represent the message passing and aggregation process as follows:

TABLE I. THE PERFORMANCE COMPERISON ON FOUR DATASETS.

| Dataset | Metric | MF | LightGCN | SGL | GTN | PDA-GNN | GraphDA | LFA-GCL |
|---|---|---|---|---|---|---|---|---|
| Yelp | Recall@20 | 0.0619 | 0.0765 | 0.0825 | 0.0806 | 0.0690 | 0.0782 | **0.0837** |
|  | NDCG@20 | 0.0485 | 0.0607 | 0.0656 | 0.0642 | 0.0557 | 0.0622 | **0.0668** |
|  | Recall@40 | 0.0906 | 0.1249 | 0.1342 | 0.1318 | 0.1141 | 0.0128 | **0.1368** |
|  | NDCG@40 | 0.0569 | 0.0784 | 0.0848 | 0.0828 | 0.0722 | 0.0816 | **0.0862** |
| Tmall | Recall@20 | 0.0368 | 0.0492 | 0.0514 | 0.0511 | 0.0431 | 0.0495 | **0.0522** |
|  | NDCG@20 | 0.0325 | 0.0436 | 0.0455 | 0.0453 | 0.0381 | 0.0439 | **0.0463** |
|  | Recall@40 | 0.0628 | 0.0794 | 0.0828 | 0.0835 | 0.0709 | 0.0799 | **0.0846** |
|  | NDCG@40 | 0.0420 | 0.0567 | 0.0585 | 0.0596 | 0.0504 | 0.0573 | **0.0605** |
| Hetrec-ML | Recall@20 | 0.0948 | 0.0989 | 0.1063 | 0.0992 | 0.1062 | 0.1147 | **0.1205** |
|  | NDCG@20 | 0.2644 | 0.3156 | 0.3295 | 0.3155 | 0.3102 | 0.3422 | **0.3653** |
|  | Recall@40 | 0.1225 | 0.1625 | 0.1731 | 0.1628 | 0.1797 | 0.1881 | **0.1936** |
|  | NDCG@40 | 0.2737 | 0.2945 | 0.3092 | 0.2950 | 0.3013 | 0.3252 | **0.3447** |
| Amazon | Recall@20 | 0.0753 | 0.0841 | 0.0977 | 0.0872 | 0.0785 | 0.0871 | **0.1026** |
|  | NDCG@20 | 0.0488 | 0.0568 | 0.0642 | 0.0582 | 0.0528 | 0.0604 | **0.0680** |
|  | Recall@40 | 0.0836 | 0.1277 | 0.1433 | 0.1332 | 0.1179 | 0.1317 | **0.1552** |
|  | NDCG@40 | 0.0523 | 0.0700 | 0.0791 | 0.0722 | 0.0648 | 0.0744 | **0.0840** |

TABLE II. STATISTICS ABOUT DATASETS

| Dataset | #Users | #Items | #Interaction | Density |
|---|---|---|---|---|
| Yelp | 29601 | 24734 | 1374594 | 0.00187 |
| Tmall | 47939 | 41390 | 2619389 | 0.00132 |
| Hetrec-ML | 2113 | 10109 | 855598 | 0.0446 |
| Amazon | 20000 | 20000 | 571994 | 0.00142 |

$$h_k^{(u)} = \sum_{l=0}^{L} \hat{R}_{k,:} \cdot E_l^{(i)} = \sum_{l=0}^{L} P_{k,:} \cdot \left(Q^T \cdot E_l^{(i)}\right),$$
$$h_j^{(i)} = \sum_{l=0}^{L} \hat{R}_{j,:}^T \cdot E_l^{(u)} = \sum_{l=0}^{L} Q_{j,:} \cdot \left(P^T \cdot E_l^{(u)}\right), \quad (10)$$

where $h_k^{(u)}$ and $h_j^{(i)}$ are the augmented representation learned for user $u_k$ and item $i_j$ from the augmented view. According to the associative law of matrix multiplication, we first compute the product of the last two matrices (e.g., $Q^T \cdot E(i)\ l$), because the dimensions $|U| \times f$ and $|I| \times f$ of $P$ and $Q$ are much smaller than $|U| \times |I|$ of $\hat{R}$, it can alleviate the computational efficiency challenge posed by dense large matrix multiplications.

### C. Graph Contrastive Learning Module

We implement a dual-channel [61, 62] graph contrastive learning by constructing an additional view. In the LFA-GCL model, recommendation with global collaboration is achieved by directly using the node representation of the main view and the augmented node representation to calculate the contrastive learning loss. We consider the representation of the same node from different views as positive node pairs (i.e., $\{(e_i, h_i)|\ i \in U\}$), while different nodes are considered as negative node pairs (i.e., $\{(e_i, h_j)|\ i, j \in U, i \neq j\}$). We compute an InfoNCE about users using the node embedding $h^{(u)}$ from the augmented view constructed using LFA and the node embedding $e^{(u)}$ from the main view as follows:

$$L_{cl}^{(u)} = \sum_{k=0}^{U} -\log \frac{\exp\left(s\left(e_k^{(u)}, h_k^{(u)}\right)/\tau\right)}{\sum_{k'=0}^{U} \exp(s(e_k^{(u)}, h_{k'}^{(u)})/\tau)}, \quad (11)$$

where the similarity function $s(\cdot)$ between two vectors here we use the cosine similarity; Similarly, we can acquire the contrastive loss $L_{cl}^{(i)}$ about the items. By summing up these two losses, an objective function $L_{cl} = L_{cl}^{(u)} + L_{cl}^{(i)}$ is obtained for the contrastive learning module.

### D. Joint Optimization Module

In this part, we adopt multi-task joint optimization of the classic BPR loss [40] and contrastive learning loss:

$$L = L_{bpr} + \lambda_1 \cdot L_{cl} + \lambda_2 \cdot \|\Theta\|_2^2, \quad (12)$$

$$L_{bpr} = \sum_{(u,i,j) \in V} -\log(\sigma(e_u^T e_i - e_u^T e_j)), \quad (13)$$

where $\lambda_1$ is the coefficient of comparative learning loss, $\Theta$ are all the parameters of the joint optimization module, and we use its L2 regularization and control the strength with $\lambda_2$. According to (13), the loss $L_{bpr}$ with a triplet input $(u, i, j)$, $e_i$ and $e_j$ represent the node representation of the positive sample that directly interacted with user $u$ and the negative sample that has not interacted with user $u$ respectively, and $\sigma(\cdot)$ represents the sigmoid function.

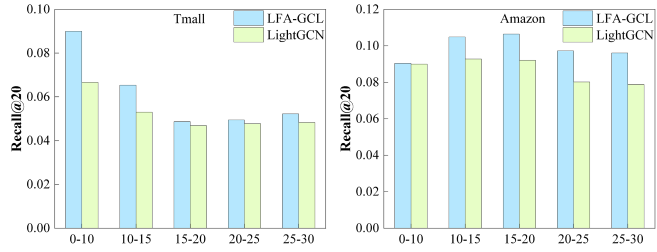

Figure 2. Performance of different user sparse interaction degree groups

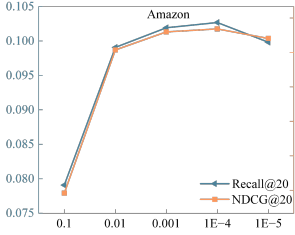
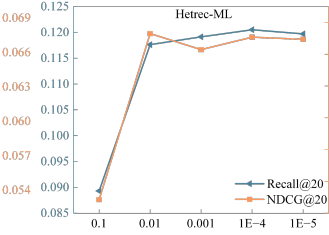

Figure 3. Impact of parameter $\lambda_1$

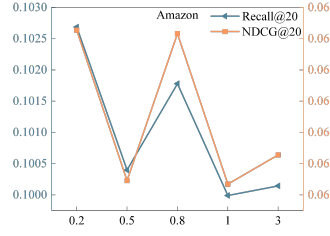
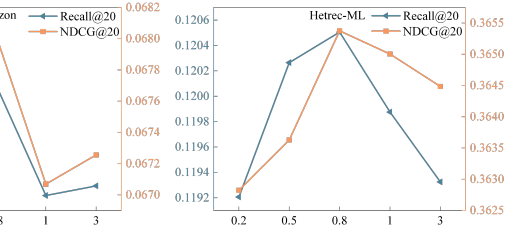

Figure 4. Impact of temperature $\tau$

## IV. EXPERIMENTS

### A. Experimental Setup

*Dataset:* We evaluate our model and the baselines using four public datasets: the Yelp dataset (rating interactions from the Yelp platform), the Tmall dataset (purchase records from the Tmall platform), the Hetrec-ML dataset (collaborative filtering data from the MovieLens platform), the Amazon dataset (interactions from the first 20,000 users and items in Amazon-book, which contains 78578 users, 77801 items, and 2240156 interactions). We divided the datasets into three sections (training set, validation set, and test set) in the ratio of 7:1:2. More of these datasets are in TABLE II.

*Evaluation Protocol:* In this paper, we use the all-ranking protocol [72] to measure the performance of Top-K recommendation, and we adopt Recall@K and NDCG@K as metrics that are widely used in existing works [17, 63] and K={20, 40}. Recall@K measures the proportion of user clicks among the K-recommended items to the entire click set. NDCG@K assesses the relevance of recommended items by considering their positions in the ranking list:

$$\mathrm{Recall}@K(u) = \frac{\left|R^K(u) \cap T(u)\right|}{\left|T(u)\right|}, \quad (14)$$

$$\mathrm{NDCG}@K = \frac{1}{|U|}\Sigma_{u \in \mathcal{U}} \frac{\sum_{n=1}^{K} \frac{I(R_n^K(u) \in T(u))}{\log(n+1)}}{\sum_{n=1}^{K} \frac{1}{\log(n+1)}}, \quad (15)$$

where $T(u)$ represents the ground-truth item set, $R^K(u)$ represents the top-$K$ recommended item set, $R_n^K(u)$ denotes the $n$-th item in the recommended list $R^K(u)$, and $I(\cdot)$ is indicator function.

*Compared Models:* We compare the proposed LFA-GCL with the following baseline methods: MF [40], LightGCN [17], SGL [63] with randomly dropping edges, GTN [80], PDA-GNN [81], and GraphDA[55].

*Parameter Settings:* To acquire the objective results, we adopt the following experimental settings:
- All comparison models are run on a GPU host computer equipped with an NVIDIA GeForce RTX 3050 OEM graphics card.
- Using Adam optimizer with batch size=2048 and learning rate=1e-3.
- Validation is performed every two iterations, and early stopping is performed after the performance on the verification set drops for 10 consecutive times.
- To provide fair comparisons, we have tuned all baselines hyperparameters within the ranges offered in their papers, but the following hyperparameters were used for all models: the embedding size=32 and all models with LightGCN as the backbone use two convolutional layers.
- For the LFA-GCL, the latent feature dimension $f$ of the LFA model is 5, the regularization weights $\lambda_1$ is tuned from {0.1, 0.01, 0.001, 0.0001, 0.00001}, and $\lambda_2$ is chosen from {1e-6, 1e-7, 1e-8}. The temperature $\tau$ is adjusted from {0.2, 0.5, 0.8, 1, 3}. The dropout rate is searched from {0, 0.1, …, 0.5}.

### B. Comparison Performance

We present a detailed comparison of metrics in TABLE I. highlighting the best baseline (underlined) and our model (bolded). Based on the data in the table, we draw the following findings:

Compared with the traditional CF method MF, all GCN-based models demonstrate superior performance. This advantage indicates that leveraging information from high-order neighbors can improve the effectiveness of representation learning.

Recommendation methods utilizing contrastive learning outperform those that do not. This indicates that InfoNCE, by emphasizing the differences between positive and negative samples, enables GCL-based methods to more effectively catch the latent user-item relationships, thereby improving the generalization ability of the model.

LightGCN is the backbone of our model and our main baseline model. Based on the results of the four datasets, our LFA-GCL model significantly improves the recommendation performance. We find that our model also outperforms SGL, indicating that our approach to augmenting graph structure based on known interaction data outperforms random data augmentation methods applied to graph structure.

*C. Resistance Against Data Sparsity*

We further validate the ability of LFA-GCL to alleviate the data sparsity through low-interaction group Recall experiments. Specifically, on the two datasets with the lowest density, we partition the data into five groups based on the interaction degree of users and calculated the average Recall@20 for each group. Fig. 2 presents the comparison of the average Recall@20 between LFA-GCL and LightGCN. Based on the results in the figure, among these users with fewer interactions, the recall performance of our model has been greatly improved compared to LightGCN. Their Recall in some groups even exceeds that of the whole dataset. Especially on the Tmall, it performs very well in user groups that are very sparse (interactions <10). This shows that our LFA-GCL can alleviate data sparsity by injecting global collaborative information.

*D. Impact of the Parameter $\lambda_1$*

The parameter $\lambda_1$ is the weight coefficient of the contrastive loss in our LFA-GCL. As shown in Fig. 3, we adjust the values of $\lambda_1$ from {0.1, …, 0.00001} on the Amazon and Hetrec-ML. The trends are generally similar across these two datasets. When $\lambda_1$=0.1, the performance on both datasets is poor. As $\lambda_1$ decreases to 0.01, Recall@20 and NDCG@20 both sharply increase. However, when $\lambda_1$=1E-5, both metrics begin to decline.

*E. Impact of the Temperature $\tau$*

The temperature parameter $\tau$ controls the negative samples discrimination level of model in the contrastive learning loss. We adjust the values of $\tau$ from {0.2, 0.5, 0.8, 1, 3} and analyze its effect on the Amazon and Hetrec-ML, and plot results as a line graph in Fig. 4. The $\tau$ varies a lot on different datasets, but when $\tau$=3, the performance is poor on both datasets. On Amazon, the performance peaks at $\tau$=0.2, while on Hetrec-ML, the performance peaks at $\tau$=0.8.

V. CONCLUSION

In this paper, we propose a dual-channel graph contrastive learning recommendation method by combining LFA and GCL. The LFA generates an augmented view to inject global collaborative signals, enhancing the ability of the model to alleviate data sparsity by comparing the main view and the augmented view. We perform experiments on four public datasets to showcase the effectiveness and advancement of the LFA-GCL model. In future work, we plan to further study data augmentation methods to simplify our model and mitigate the confounding effects of data augmentation while ensuring recommendation performance.